\documentclass[sigconf]{acmart}

\AtBeginDocument{%
  }

\copyrightyear{2026}
\acmYear{2026}
\setcopyright{cc}
\setcctype{by}
\acmConference[ICAIF '26]{7th ACM International Conference on AI in Finance}{November 14--17, 2026}{Milan, AA, Italy}
\acmBooktitle{7th ACM International Conference on AI in Finance (ICAIF '26), November 14--17, 2026, Milan, AA, Italy}
\acmDOI{10.1145/3858063.3858908}
\acmISBN{979-8-4007-3087-0/2026/11}

\newcommand{\ci}[2]{{\scriptsize[#1,\,#2]}}
\newcommand{\cis}[2]{{\scriptsize[#1,\,#2]}}
\begin{document}

\title{Adversarial Post-Training of Foundation Models for High-Frequency Trading}

\author{Alexandre Bismuth}
\orcid{0009-0008-6558-2781}
\authornote{Equal contribution.}
\email{alexandre.bismuth@cs.ox.ac.uk}
\affiliation{%
  \department{British Open-ended Learning and Discovery Lab (BOLD)}
  \department{Department of Computer Science}
  \institution{University of Oxford}
  \city{Oxford}
  \country{United Kingdom}
}
\affiliation{%
  \institution{Harvard University}
  \city{Cambridge}
  \state{MA}
  \country{United States}
}

\author{Kang Li}
\orcid{0009-0007-7847-9292}
\email{kang.li@stats.ox.ac.uk}
\authornotemark[1]
\affiliation{%
  \department{British Open-ended Learning and Discovery Lab (BOLD)}
  \department{Department of Statistics}
  \institution{University of Oxford}
  \city{Oxford}
  \country{United Kingdom}
}

\author{Aramis Marti-Shahandeh}
\orcid{0009-0007-4757-4844}
\email{aramis@robots.ox.ac.uk}
\affiliation{%
  \department{British Open-ended Learning and Discovery Lab (BOLD)}
  \department{Department of Engineering Science}
  \institution{University of Oxford}
  \city{Oxford}
  \country{United Kingdom}
}

\author{Valentin Mohl}
\orcid{0000-0002-6166-2222}
\email{valentin.mohl@cs.ox.ac.uk}
\affiliation{%
  \department{British Open-ended Learning and Discovery Lab (BOLD)}
  \department{Department of Computer Science}
  \institution{University of Oxford}
  \city{Oxford}
  \country{United Kingdom}
}

\author{Jakob Foerster}
\orcid{0000-0001-9688-2498}
\email{jakob@robots.ox.ac.uk}
\affiliation{%
  \department{British Open-ended Learning and Discovery Lab (BOLD)}
  \department{Department of Engineering Science}
  \institution{University of Oxford}
  \city{Oxford}
  \country{United Kingdom}
}

\author{Anisoara Calinescu}
\orcid{0000-0003-2082-734X}
\email{ani.calinescu@cs.ox.ac.uk}
\affiliation{%
  \department{British Open-ended Learning and Discovery Lab (BOLD)}
  \department{Department of Computer Science}
  \institution{University of Oxford}
  \city{Oxford}
  \country{United Kingdom}
}

\renewcommand{\shortauthors}{Bismuth et al.}

\begin{abstract}
  While autoregressive foundation models trained on limit order book (LOB) data show state-of-the-art performance for market simulation, pre-training with cross-entropy neither penalizes the compounding of sampling errors over free-running rollouts nor targets desired market properties. We introduce a post-training method for LOB foundation models using a Wasserstein generative adversarial network (GAN) objective. A critic, trained online on rollouts through the true, non-differentiable matching engine, guides updates from evolution strategies over rank-4 LoRA perturbations of the pre-trained generator, under a KL trust region. For efficiency, we use EGGROLL (Evolution Guided GeneRal Optimisation via Low-rank Learning), a memory-efficient variant of evolution strategies. On a held-out panel, adversarial post-training improves LOB-Bench realism by $17.6\%$ (WS-21) and $17.0\%$ (Mean-$L_1$), while selection-matched random-perturbation and compute-matched continued-pre-training controls yield no improvement. Post-training is thus a necessary stage that more pre-training does not replace. Per-position divergence measurements show that it slows error accumulation over 500-message rollouts from $3.1\times$ to $2.3\times$. Adversarial post-training is also inexpensive, requiring under $50$\,M training tokens and $12$ GPU-hours per seed, three orders of magnitude below the $48$\,B-token pre-training budget. Ablations over critic inputs, optimizers, and the population--rollout trade-off, together with an analysis of reward hacking, yield practical design guidelines for adversarial post-training.
\end{abstract}

\ccsdesc[500]{Computing methodologies~Neural networks}
\ccsdesc[300]{Computing methodologies~Continuous space search}
\ccsdesc[500]{Applied computing~Economics}

\keywords{Foundation models, Limit order books, Market simulation, Evolution strategies, Post-training, Adversarial networks}

\maketitle

\begin{figure*}[!t]
    \includegraphics[width=\textwidth]{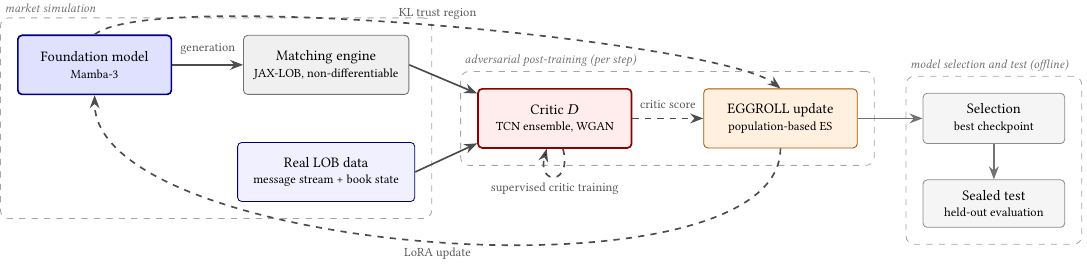}
    \Description{Three-panel diagram. Left: a model continues order-book messages through an exchange simulator. Center: a critic scores generated rollouts against real ones and EGGROLL turns the scores into low-rank weight updates. Right: checkpoints are selected with LOB-Bench and scored on held-out data.}
    \caption{\textbf{Market simulation (left):} a model continues LOB messages through a non-differentiable exchange simulator. \textbf{Adversarial post-training (center):} a critic trained online scores generated rollouts against real message streams, and EGGROLL converts scores into rank-4 LoRA~\cite{hu2022lora} perturbations under a KL trust region. \textbf{Evaluation (right):} offline selection with LOB-Bench and held-out evaluation to mitigate reward hacking. Solid arrows carry data; dashed arrows carry learning signals.}
    \label{fig:post_training_pipeline}
\end{figure*}

\section{Introduction}
Recent work has established foundation models for high-frequency trading as the state-of-the-art for market simulation through autoregressive training on limit order book (LOB) messages, executed through a matching engine~\cite{nagy2023generative,wheeler2024marketgpt,li2025mars}. These models are pre-trained with cross-entropy, maximizing next-token likelihood over historical message streams and thus implicitly matching the market's conditional next-message distribution. However, next-token prediction leaves a gap that more pre-training does not close: it creates compounding errors over long horizons, an effect known as exposure bias~\cite{bengio2015scheduled,ranzato2016sequence}, slowly drifting to unrealistic scenarios.\par\vspace{\baselineskip}

Here, adversarial methods offer a promising solution: train a critic to distinguish real LOB sequences from generated rollouts, and update the pre-trained model to fool it. This method implicitly estimates a divergence between the two sequence distributions and, unlike hand-crafted statistics, continually reallocates capacity to whichever discrepancies the generator currently exhibits. The obstacle is that rollouts are executed through the exchange's matching engine, which is non-differentiable: no gradient connects the critic's score back to the generator's parameters. We therefore turn to zero-order optimization---methods such as EGGROLL~\cite{eggroll} or GRPO~\cite{shao2024deepseekmath} that require only scalar rewards evaluated on rollouts, never a backward pass through the simulator.\footnote{Source code: \url{https://github.com/alexandre-bismuth/WGAN-Post-Training}.}

Furthermore, while this work focuses on market realism for LOB foundation models, our framework can be adapted to any desired metric. One promising target is market-impact fidelity, relevant to agentic trading (Section~\ref{sec:generalization}). Outside of the HFT literature, our approach also applies to other autoregressive models where compounding errors create long-horizon drifts and non-differentiability prevents the use of gradient-based methods.

\subsection{Contributions}
\vspace{0.5em}
\begin{enumerate}
    \item \textbf{An adversarial post-training framework for HFT foundation models.} A Wasserstein critic, trained online and from scratch on free-running rollouts executed through the true, non-differentiable matching engine, steers gradient-free optimizers that update a pre-trained generator through rank-4 LoRA perturbations under a KL trust region. \vspace{0.5em}
    \item \textbf{Improved market realism at negligible cost.} Tested over held-out months, post-training improves WS-21 by $17.6\%$ and Mean-$L_1$ by $17.0\%$ while consuming under $50$\,M training tokens and $12$ GPU-hours---a significant refinement after $170$ GPU-hours and ${\approx}48$\,B tokens of pre-training.  \vspace{0.5em}
    \item \textbf{An anatomy of adversarial post-training.} Ablation experiments allow us to define robust guidelines for pipeline design: raw data roughly doubles the gains of hand-designed inputs, small populations lose nothing at fixed compute, and reward-hacking requires careful evaluation protocols.
\end{enumerate}

\section{Background}

\paragraph{Foundation models for market simulation}

\begin{figure}[h]
    \centering
    \includegraphics[width=\columnwidth]{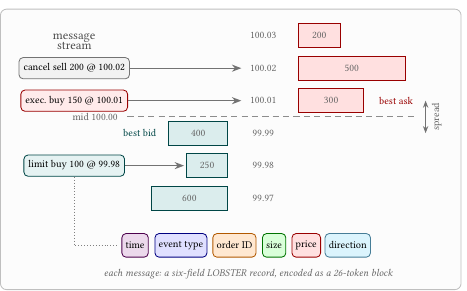}
    \Description{A limit order book drawn as bid and ask volume bars at each price level. Example messages (a cancellation, an execution and a new limit order) point to the levels they change, and a legend lists the six fields of a LOBSTER record.}
    \caption{The limit order book and its message stream from records in LOBSTER format. Bar length represents volume.}
    \label{fig:concept}
\end{figure}

Electronic markets provide an ideal starting point for the development of foundation models thanks to the vast quantity of raw data. Every order submission, cancellation, deletion, and execution is a time-stamped message, and the exchange's matching engine deterministically folds the message stream into the limit order book (LOB, Figure~\ref{fig:concept}). Data feeds such as LOBSTER~\cite{huang2011lobster} preserve this stream losslessly, turning market simulation into a sequence-modeling problem: an autoregressive model over tokenized messages, pre-trained with cross-entropy on historical flow and combined with a matching engine, defines a fully interactive synthetic market~\cite{nagy2023generative,li2025mars}. GPU-native engines make such closed-loop rollouts fast enough to train and evaluate against~\cite{frey2023jaxlob}. Because next-message likelihood alone does not certify realism, generated markets are assessed distributionally over longer sequences: LOB-Bench~\cite{nagy2025lobbench} scores the discrepancy between real and generated rollouts across microstructure statistics (e.g. spreads, depths, volumes, order-flow imbalance, inter-arrival times) and is the evaluation standard we adopt.

\paragraph{Generative Adversarial Networks}

A Generative Adversarial Network (GAN) trains a generator against a discriminator that learns to separate real from generated samples~\cite{goodfellow2014generative}. In the Wasserstein formulation, a 1-Lipschitz critic estimates the Wasserstein-1 distance between the two distributions via Kantorovich--Rubinstein duality, yielding a training signal that stays informative even when the two distributions are nearly disjoint~\cite{arjovsky2017wasserstein}; the Lipschitz constraint is enforced by gradient penalties~\cite{gulrajani2017improved} or spectral normalization~\cite{miyato2018spectral}, and training is stabilized by two-time-scale updates~\cite{heusel2017ttur} and the zero-centered $R_1$ penalty, which restores local convergence~\cite{mescheder2018convergence}.

\paragraph{Zero-order optimization methods}
Zero-order methods optimize objectives that can be evaluated but not differentiated: instead of a gradient, they build an update from the scores of sampled candidates~\cite{spall1992multivariate,wierstra2014natural}. On one hand, policy-gradient methods assign each sampled sequence an advantage, defined as how much better it scored than a baseline, and raise or lower its log-likelihood accordingly, shifting probability mass toward higher-reward samples. GRPO~\cite{shao2024deepseekmath} replaces the learned value baseline with the group itself: it draws a group of rollouts per context, z-scores their rewards into group-relative advantages, and constrains the update with a KL penalty to a reference model. On the other hand, Evolution Strategies (ES) sample perturbations of the parameters, score each perturbed model on the objective, and combine the perturbations, weighted by rank-shaped antithetic scores, into an update~\cite{salimans2017evolution}. EGGROLL~\cite{eggroll} makes this efficient by drawing each perturbation as a rank-$r$ outer product (collapsing per-member memory from $O(mn)$ to $O(r(m+n))$), and is competitive with other policy optimization methods used to fine-tune modern large language models. Both families need only scalar rollout scores and therefore tolerate a non-differentiable environment; they differ in where the update is estimated (weight space versus token space) and in what must be differentiated (nothing versus the generator's own log-probabilities).

\section{Related Work}

\paragraph{Data-driven market simulation}
Agent-based simulators such as ABIDES~\cite{byrd2020abides} compose hand-crafted trading agents around an exchange model; they are interactive by construction but struggle to match the fine structure of real order flow. A data-driven line therefore replaces the agent pool with a learned world agent: Stock-GAN generates order streams with a conditional Wasserstein GAN, sidestepping the non-differentiable exchange by learning a neural approximation of the matching mechanism~\cite{li2020generating}; Coletta et al.\ develop CGAN world agents deployed inside multi-agent simulations~\cite{coletta2021towards,coletta2022learning} and analyze their explainability and robustness~\cite{coletta2023conditional}; Hultin et al.\ factorize message generation into a chain of conditional distributions~\cite{hultin2023generative}; and TRADES brings denoising diffusion to the same world-agent setting~\cite{berti2025trades}. Most recently, token-level autoregressive foundation models~\cite{nagy2023generative,wheeler2024marketgpt,li2025mars} have become the strongest generative simulators. Our work is orthogonal to this scaling approach: we take a pre-trained foundation model as given and ask how a second, adversarial training stage improves it.

\paragraph{Adversarial training for sequence models}
For discrete sequences, discriminator gradients cannot flow through sampling; SeqGAN resolved this by treating generation as a policy updated with REINFORCE against discriminator rewards~\cite{yu2017seqgan}, a pattern our GRPO arm modernizes. Closest to us in spirit, adversarial post-training refines pre-trained diffusion models against real data after likelihood-style pre-training~\cite{lin2025diffusion}: pre-train for coverage, then optimize an adversarial objective for fidelity. We transport this two-stage pattern to market generation, where the adversarial stage must additionally traverse a hard exchange simulator, and pair it with gradient-free optimizers rather than backpropagation.

\paragraph{Reward-driven post-training}
Post-training pre-trained models against scalar rewards is standard practice in the LLM literature, via policy-gradient methods such as GRPO~\cite{shao2024deepseekmath} or, increasingly, evolution strategies~\cite{salimans2017evolution,qiu2025evolution}. Two recent observations shape our design point. First, while hyperscale ES pre-training relies on enormous populations~\cite{eggroll}, the neighborhood of a large pre-trained model is dense with improving directions~\cite{gan2026neural}, suggesting that post-training needs orders of magnitude fewer. Second, any learned reward is eventually over-optimized~\cite{gao2023scaling}; our online critic inherits this in the form of a measurable hacking window. Unlike RLHF-style pipelines whose reward model is frozen, our reward is a live adversary trained online, exploited on a documented schedule, and disciplined by external selection on held-out statistics.

\section{Methods}
\label{sec:methods}

\subsection{Preprocessing}

\paragraph{Dataset} The training corpus contains GOOG stock information from 2022 to 2025, covering about 1,000 trading days. Testing is completed over 28 trading days in January and February 2026. The pre-trained generator (the anchor in our study) learns from $\approx50\%$ of the training corpus. Post-training is performed on unseen training data, avoiding any unfair specialization or recency advantage. Unlike existing literature, we train on the full message-stream instead of focusing on the top-k levels of the books. 

The test period shows no dramatic regime shift. Event composition and volatility are stable across periods, while activity and price trend upward through the training years (Table~\ref{tab:data_stats}).

\begin{table}[h]
  \captionsetup{singlelinecheck=false}
  \caption{Statistics for the GOOG stock.}
  \vspace{-1em}
  \label{tab:data_stats}
  \small
  \setlength{\tabcolsep}{2.5pt}
  \begin{tabular}{@{}lrrrrr@{}}
    \toprule
    Period & Msgs/day (M) & Exec. (\%) & Spread (bps) & Daily RV (\%) & Mid (\$) \\
    \midrule
    2022--2025 & 4.41 & 1.45 & 1.18 & 1.44 & 157 \\
    2025 Q1 & 5.95 & 1.20 & 1.25 & 1.50 & 183 \\
    2025 Q2 & 6.87 & 1.55 & 1.20 & 1.80 & 165 \\
    2025 Q3 & 5.15 & 1.40 & 1.14 & 1.31 & 210 \\
    2025 Q4 & 8.03 & 0.88 & 1.48 & 1.56 & 286 \\
    Test    & 6.17 & 1.29 & 1.24 & 1.70 & 329 \\
    \bottomrule
  \end{tabular}
\end{table}

\paragraph{Limit order book} The LOBSTER~\cite{huang2011lobster} format contains message-level information about the GOOG stock. It provides event types, $10^{-4}$ USD prices, volume, IDs, and nanosecond-resolution timestamps. We discard out-of-hours messages, hidden executions, cross trades, trading-halt indicators, and clip volumes at 9,999 shares. 

\paragraph{Tokenization} LOBSTER messages are each encoded as 26 tokens over a shared vocabulary of $|\mathcal{V}| = 2{,}112$, with four reserved special tokens (mask, hidden-event, not-applicable, start-of-sequence). Fields cover the message's event type, direction, timing, price, and size, plus the referenced order's attributes where applicable; sizes and price magnitudes are split into digit-token pairs (base-100 and base-1{,}000 respectively), shrinking the vocabulary while covering $\pm 999{,}999$ ticks. Absolute timestamps are kept in the loss for temporal consistency but are filled deterministically at inference. Logit masking enforces each position's valid-token set, guaranteeing syntactic validity (e.g., event types in $\{1,2,3,4\}$) though not semantic validity (e.g., canceling an order absent from the book). 

\paragraph{Generator} The pre-trained generator is a $78.5$\,M-parameter Mamba-3 state-space model~\cite{lahoti2026mamba}. Two message-encoder layers embed the token stream and two book-encoder layers embed the order-book state; both feed six fused SSM layers (width $d_{\text{model}}=1{,}024$, state size $128$) and a linear head over $\mathcal{V}$. The model conditions on $500$-message contexts ($13{,}000$ tokens) and generates by sampling each token from its $50$ most likely values (top-$k$ sampling, $k{=}50$), at about $50$k tokens per second on a single GH200 GPU.

\begin{figure}[b]
    \centering
    \includegraphics{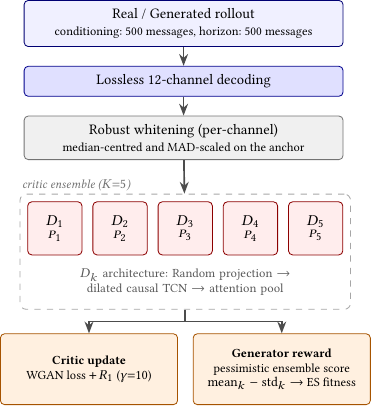}
    \Description{Each tokenized message is decoded into twelve numeric channels, normalized, and passed to an ensemble of convolutional critics whose pessimistic score is sent to the generator.}
    \caption{Raw-message critic. Each 26-token message is decoded into 12 channels while absolute values and IDs are withheld as trivial tells. Channels are median-centered and scaled by $1.4826\cdot\mathrm{MAD}$ (median absolute deviation) to match a Gaussian distribution. Spectral-normalized TCN members, each behind its own fixed random projection, score the 500-message continuation and the generator receives the pessimistic ensemble score.}
    \label{fig:raw_critic}
\end{figure}

\paragraph{Critic input} Generated data is fed into the critic to promote representation learning. We perform lossless transforms into channels (Figure~\ref{fig:raw_critic}): four one-hot fields for event type, a direction, a price, a size, a time difference, and four reference channels, defined for messages that reference another order. We intentionally discard absolute price, absolute time, and order IDs as trivial tells can separate real from fake without reflecting realism: NA in fake rollouts, non-stationary or pure bookkeeping in real data. These channels are combined with median normalization to mitigate heavy tails that dominate scales. The pool of 512 real context windows is redrawn every five steps, keeping the critic's real set briefly stationary while preventing memorization of individual windows. At each step, $Q{=}128$ contexts are sampled without replacement and shared across perturbations
so fitness circumvents context-sampling noise.

\subsection{Adversarial training}

\paragraph{Wasserstein GAN (W-GAN)} At each step, the generator rolls out $G{\times}Q = 16{\times}128 = 2{,}048$ perturbed continuations; the critic scores each rollout, and we apply EGGROLL to the generator's interior projection kernels. The critic is trained from scratch and updated online alongside the generator on the W-GAN objective with $R_1$ penalty ($\gamma{=}10$), averaged over the $K{=}5$ independent ensemble members, using Adam at a learning rate of $3{\times}10^{-4}$ with $\beta_1{=}0$~\cite{heusel2017ttur}. The $R_1$ term is a zero-centered gradient penalty on real windows only, $\tfrac{\gamma}{2}\,\mathbb{E}_{x\sim p_{\mathrm{real}}}\!\left[\lVert\nabla_x D(x)\rVert^2\right]$, which flattens the critic around the data manifold and keeps the online adversarial updates locally convergent~\cite{mescheder2018convergence}. Trained in this way, its decision boundary tracks the live discrepancies that free rollouts exhibit, including compounding errors that cross-entropy never penalizes.

\paragraph{EGGROLL} Since no gradient crosses the matching engine, the generator is updated with EGGROLL~\cite{eggroll}: $G{=}16$ members (eight antithetic directions) of rank $r{=}4$ with $\sigma{=}0.003$ (swept), evaluated on $Q{=}128$ shared contexts. A direction's fitness is its pessimistic ensemble score ($\mathrm{mean}_k - \mathrm{std}_k$), penalized by the KL trust region (below), centered-ranked per context, and averaged over contexts; the update is applied by Adam at $5{\times}10^{-4}$ learning rate. For comparison, we swap the ES update for GRPO~\cite{shao2024deepseekmath} under the same critic and contexts, with per-context groups ($G{=}8$), z-scored advantages, a single on-policy step, and a stronger trust region ($\lambda=0.3$), as raw advantages lack the rank transform's bounding.

\paragraph{Generator updates} The pre-trained generator is frozen except for the interior projection kernels of the SSM blocks, which EGGROLL perturbs with rank-4 LoRA factors and updates in full. We intentionally exclude other parts of the SSM architecture to keep ES low-variance while preserving the pre-trained dynamics encoded in the SSM core, embeddings, and decoder head. On top of that, a Kullback-Leibler divergence trust region of $\lambda = 0.1$ is applied inside the fitness to prevent reward-hacking drift.

\subsection{Evaluation}

\paragraph{Metrics} Rollout realism is measured with LOB-Bench~\cite{nagy2025lobbench}, the standard evaluation framework for generative message-level LOB models: real and generated continuations are mapped through a fixed battery of scoring functions covering microstructure statistics such as the bid--ask spread, returns, order-book volumes and imbalance, and message inter-arrival times. We aggregate the benchmark's 21 unconditional scores into two summary numbers: \textbf{WS-21}, the mean over the 21 scores of the Wasserstein-1 distance between real and generated marginals, and \textbf{Mean-$L_1$}, the mean of the corresponding $L_1$ histogram distances. The $L_1$ histogram distance reveals localized density mismatches, while the Wasserstein-1 distance responds to global distributional shape, including tails. Uncertainty is quantified with a paired window-level block bootstrap: whole evaluation windows are resampled with replacement ($500$ replicates), each replicate re-scores real and generated continuations from the same drawn windows, and model-versus-anchor deltas are formed replicate-wise from these shared draws, so window-composition noise cancels in the comparison. Day-clustered resampling over the 28 panel days yields consistent intervals.

\paragraph{Selection} Any learned reward is eventually over-optimized~\cite{gao2023scaling}, and ours is a live adversary: past a certain point the generator raises its critic score by exploiting the critic rather than by becoming more realistic, so the last checkpoint cannot be the deployed one. A three-seed pilot trained to 50 steps locates that turn (Section~\ref{sec:hacking}): realism improves monotonically through steps $\approx$15--20, then decays. We therefore train the headline seeds for 30 steps and score a checkpoint grid spaced every two steps. Each is scored offline with WS-21 on 2,048 context windows (1,024 for comparative studies) drawn from the training timeframe, and the best is selected as the winner; the picks land at steps 10--20 (median 16). The criterion uses no critic signal by design, so a checkpoint cannot select itself through the same exploit that inflated its reward.

\paragraph{Sealed test} Selected checkpoints are scored once on 4,096 disjoint windows (2,048 windows for comparative studies) from a 28-trading-day panel of most recent unseen data (January--February 2026). Results are compared against the pre-trained model re-scored on the same windows.

\paragraph{Control} We compare against a selection-matched null in which training is replaced by drawing $N{=}8$ random LoRA perturbations of the base model through seed, matched to the deployed checkpoints in parameter norm, and passed the identical selection and sealed-test pipeline. This process bounds how much apparent improvement the pipeline can manufacture from selection bias.

\section{Results}

The optimized pipeline results are shown in Table~\ref{tab:sealed}: adversarial post-training improves held-out realism for every seed, with WS-21 falling by $17.6\%$ against the base pre-trained model and Mean-$L_1$ by $17.0\%$; the paired $95\%$ bootstrap interval excludes zero for the ten-seed mean and for each seed individually (per-seed WS-21 range $-11.3\%$ to $-23.9\%$), while the selection null ends above the anchor and the IsoFLOP control within noise of the pre-trained model.

\begin{table}[h]
  \caption{Sealed test at $4{,}096$ held-out windows (28-day panel). Each $\Delta$ is paired against the frozen pre-trained generator on the same windows, with window-block bootstrap $95\%$ CIs; the post-trained $\Delta$ is the mean over $n{=}10$ seeds.}
  \label{tab:sealed}
  \setlength{\tabcolsep}{2.4pt}
  \begin{tabular}{@{}lll@{}}
    \toprule
     Model & $\Delta$ WS-21 [95\% CI] & $\Delta$ Mean $L_1$ [95\% CI] \\
    \midrule
    Pre-Trained & --- & --- \\
    Post-Trained & $\mathbf{-17.6\%}$\,\ci{-19.7}{-15.3} & $\mathbf{-17.0\%}$\,\ci{-18.9}{-15.8} \\
    Selection-null & $+5.3\%$\,\ci{+2.5}{+8.1} & $+7.8\%$\,\ci{+5.9}{+9.6} \\
    IsoFLOP control & $+0.5\%$\,\ci{-2.3}{+3.4} & $-0.4\%$\,\ci{-2.2}{+1.4} \\
    \bottomrule
  \end{tabular}
\end{table}

This adversarial pipeline consumes only about $50$\,M training tokens---a $10^3\times$ reduction from the ${\approx}48$\,B-token pre-training corpus. With 12 GPU-hours per seed, our pipeline shows that HFT foundation models can largely benefit from adversarial post-training as an inexpensive refinement stage.

The gains also reach the failure mode that motivates post-training: error compounding over free-running rollouts. Measured position by position against the true continuation, the anchor's divergence grows $3.1\times$ over $500$ generated messages; averaged over the ten sealed seeds, post-training slows the accumulation to $2.3\times$, accumulating $26\%$ less excess divergence (paired day-bootstrap $95\%$ CI $[-33\%,-20\%]$), with every seed accumulating less than the anchor. Neither control reproduces the reduction---the selection-matched null accumulates $27\%$ more than the anchor, while continued pre-training, at the IsoFLOP budget of the full experiment, accumulates at the anchor's rate (Figure~\ref{fig:compounding}). Section~\ref{sec:compounding} details the estimator, the controls, and their interpretation.

\subsection{Training dynamics}

\subsubsection{Reward hacking}
\label{sec:hacking}

The per-seed training dynamics also reveal the need for external selection: with three seeds trained on 50 steps, all seeds improve monotonically to steps $\approx$15--20, reaching $-16.4\%$ to $-21.7\%$ against the anchor on the selection windows, then decay as the generator begins to exploit the critic---seed 1 returns past the anchor ($+25.4\%$) by step 50 and seeds 0 and 2 to roughly anchor level---and the WS-21 grid selection catches the turn in all three seeds. The softer sealed-test deltas ($-17.5\%$ mean, against $-19.4\%$ at selection) reflect ordinary selection optimism; the sign is preserved out of sample.\par\vspace{\baselineskip}

The critic tells the same story from the other side. Its held-out real-versus-fake discrimination starts strong (AUC ${\approx}0.62$ at the first step) and declines steadily to chance level (${\approx}0.51$, calibration cross-entropy peaking near $0.80$ nats) at steps 20--25: the generator genuinely closes the gap the critic can measure, exactly where grid selection places the turn. Past the turn the equilibrium destabilizes: discrimination recovers to AUC ${\approx}0.64$ and the cross-entropy drops below the $\ln 2$ chance level for the only time in training, the online critic cleanly re-separating an adversary whose continued reward ascent has decoupled from realism. The ten 30-step headline seeds retrace the approach leg of the same arc (mean AUC $0.62 \to 0.52$ by step 20).\par\vspace{\baselineskip}

\subsubsection{Loss behavior}
\label{sec:loss}

Adversarial gains cost little likelihood. Teacher-forced next-token cross-entropy on $512$ held-out real windows rises by ${+}0.10\%$ ($95\%$ CI $[+0.07\%,+0.12\%]$) by step 20 across the ten headline seeds, where it stays through step 30. Nor is the concession a uniform drift: per-seed values at step 30 span $-0.38\%$ to $+0.41\%$, while the KL divergence to the anchor grows monotonically in every seed ($2.0 \to 6.2 {\times} 10^{-3}$ between steps 10 and 30). The update thus moves steadily away from the pre-trained model while leaving its likelihood almost untouched. 

\subsection{Compute efficiency}

All of our results are obtained with populations of just $G{=}16$ members (eight antithetic direction pairs), five orders of magnitude below the ${\sim}1$M-member populations EGGROLL was introduced with for from-scratch pre-training~\cite{eggroll}. The regimes differ fundamentally. Post-training explores the immediate neighborhood of a strong anchor. The scarce resource is instead the reliability of each direction's score, since market rollouts are a noisy fitness signal. We therefore average each direction's fitness over $Q{=}128$ shared contexts, spending the rollout budget on update stability rather than population size. At a fixed budget $G{\times}Q$, the outcome should then be governed by the total rather than the direction count, and small populations should lose nothing, as Section~\ref{sec:frontier} confirms experimentally.

\subsubsection{Compute-optimal frontier}
\label{sec:frontier}

We ablate the split between direction count $G$ and contexts per direction $Q$ at a fixed budget of $2{,}048$ rollouts per step, three seeds per arm, holding everything else constant. On the sealed test, different allocations all sit well inside the within-arm seed spread. The split shows instead at selection, where larger-$G$ arms tend to select earlier checkpoints (median picked step 20 at $G{=}16$ versus 15 at $G{=}32$ and $G{=}64$)---sharper per-step gradient estimates climb the critic's proxy faster and approach the reward-hacking window sooner, while the co-trained critic receives the same number of updates to keep pace.\par\vspace{\baselineskip}

Restoring $Q{=}128$ at $G{=}64$ quadruples the budget to $8{,}192$ rollouts per step and buys nothing: its seed lands at $-13.5\%$ WS-21 and $-15.8\%$ Mean-$L_1$, below the WS-21 seed range of the fixed-budget arms. Two conclusions follow. At a fixed budget, the sealed outcome is insensitive to how rollouts are split between directions and contexts. Furthermore, growing the population without shrinking the contexts per direction (to more closely resemble ES pre-training) buys no additional sealed realism either: against a critic trained online, the binding constraint is the critic's robustness, not population size.

\subsubsection{Benefits over pre-training}

The improvement is not a token-budget effect that more pre-training could reproduce. As a control, we continue pre-training the anchor on the same unseen corpus under its original recipe, with compute matched to a single post-training seed's full experiment and a fresh cosine learning-rate schedule spanning that budget. Fully annealed, the control lands within noise of the pre-trained generator ($+0.5\%$ WS-21 and $-0.4\%$ Mean-$L_1$, paired CIs spanning zero; Table~\ref{tab:sealed}). Additional likelihood training on the same corpus therefore reproduces none of the adversarial gain, revealing the value of the critic's sequence-level signal beyond data volume or extra compute. This benefit is also confirmed over the compounding error analysis of Section~\ref{sec:compounding}, where extra pre-training does not yield improvements over the pre-trained generator.

\subsection{Critic and optimizer ablations}
\label{sec:critic_ablations}

Holding the training loop fixed, we ablate the two central design choices of the pipeline: what the critic observes, and how its scores become generator updates.

\subsubsection{Input features}

The raw-message critic of Section~\ref{sec:methods} is the endpoint of an ablation over the critic's input. Reusing the generator itself, scoring rollouts through the frozen backbone's pooled hidden state with a shallow head, appears free but is structurally exploitable: the adversarial reward turns anti-aligned with held-out realism (correlation $+0.5$--$0.8$ between reward and LOB-Bench error), and two of three seeds decay back to the anchor. Hand-designed per-step microstructure descriptors, scored by the same spectral-normalized TCN ensemble, restore a reliable signal but restrict the critic to discrepancies the descriptors can express. The losslessly decoded message stream removes both failure modes, leaving representation learning to the adversarial objective.\par\vspace{\baselineskip}

Table~\ref{tab:critic_input} quantifies the comparison under the otherwise-identical protocol: three seeds per arm, $G{\times}Q=16{\times}128$, 50 steps, unseen-corpus contexts, identical selection, and one single-program sealed test in which every row is generated under common random numbers and scored against the same anchor. The raw-message critic roughly doubles the descriptor critic's sealed effect, improving WS-21 by $17.0/15.7/19.7\%$ per seed (Mean-$L_1$ $17.2/18.4/21.1\%$) against a descriptor-critic mean of $8.5\%$ ($6.7\%$): every raw-critic seed beats every descriptor-critic seed on both metrics, and the descriptor arm's weakest seed is statistically indistinguishable from the anchor ($-2.4\%$ WS-21, paired CI spanning zero). The backbone critic's selected checkpoints score well on average but are the least reliable of the three arms, spanning $-11.5\%$ to $-28.2\%$ across seeds---consistent with an exploitable critic rather than the reward remaining a faithful realism signal throughout training.

\begin{table}[t]
  \caption{Critic-input ablation. Single-program sealed test at $2{,}048$ held-out windows (28-day panel), all rows sharing one anchor; $\Delta$ = mean paired change vs.\ that anchor across three seeds, with the per-seed range in brackets.}
  \label{tab:critic_input}
  \setlength{\tabcolsep}{2pt}
  \begin{tabular}{@{}lrr@{}}
    \toprule
    Critic input & $\Delta$WS-21 $\downarrow$ & $\Delta$Mean $L_1$ $\downarrow$ \\
    \midrule
    Generator backbone & $-21.4\%$ {\scriptsize$[-11.5,-28.2]$} & $-18.7\%$ {\scriptsize$[-10.0,-23.7]$} \\
    Microstructure descriptors & $-8.5\%$ {\scriptsize$[-2.4,-11.6]$} & $-6.7\%$ {\scriptsize$[+0.2,-13.3]$} \\
    Raw messages & $-17.5\%$ {\scriptsize$[-15.7,-19.7]$} & $-18.9\%$ {\scriptsize$[-17.2,-21.1]$} \\
    \bottomrule
  \end{tabular}
\end{table}

\subsubsection{Optimization method}
\label{sec:optimizer}

The critic's signal is not specific to evolution strategies. As a comparison, we swap the ES update for GRPO under an otherwise-identical pipeline and repeat the full headline protocol: ten seeds, identical in-training-timeframe selection over the checkpoint grid, and one sealed scoring at $4{,}096$ held-out windows with a norm-matched selection null of its own. Across ten seeds, GRPO shifts WS-21 by $-16.0\%$ and Mean-$L_1$ by $-17.3\%$ against the same anchor (its selection null: $+1.8\%$, not significant), compared with $-17.6\%$ and $-17.0\%$ for EGGROLL: the adversarial reward is optimizer-agnostic. We retain EGGROLL for the headline pipeline for two reasons. First, GRPO's selection curves are jagged---sharp step-to-step swings rather than EGGROLL's smooth rise into the hacking window---making results less predictable and harder to analyze. Second, the per-position decomposition of Section~\ref{sec:compounding} shows contrasting results. GRPO slows accumulation even more than EGGROLL, but buys it with a large level offset: its marginals start $75\%$ above the anchor's in the first 25-message bin, and the slower growth only wins that offset back, leaving its total divergence at the 500-message horizon almost indistinguishable from the anchor ($-1.0\%$) and above EGGROLL's total divergence ($-11.3\%$).

\section{Discussion}

\subsection{Compounding error}
\label{sec:compounding}

\begin{figure}[h]
    \centering
    \includegraphics[width=\linewidth]{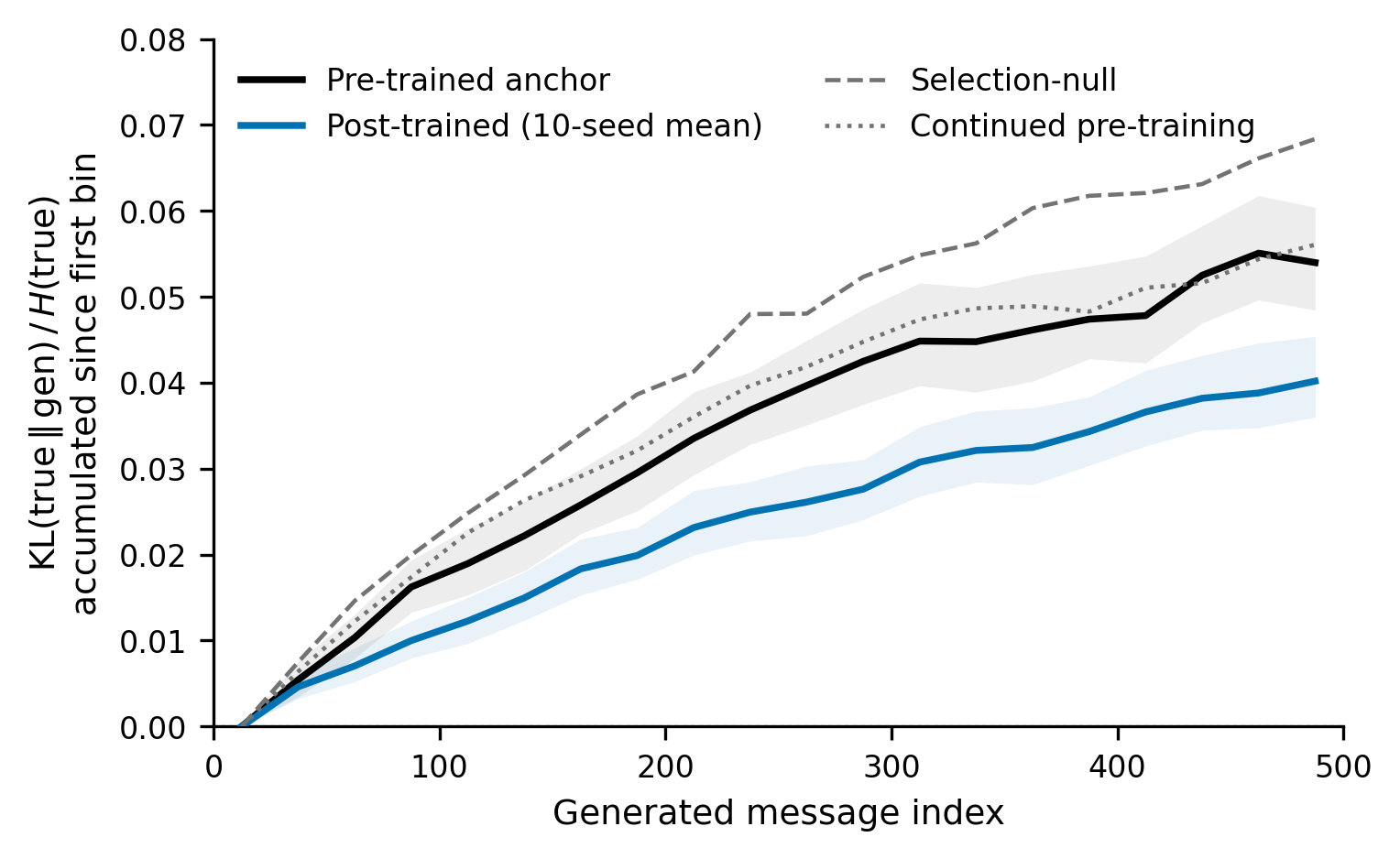}
    \Description{Line plot of accumulated divergence against position in a 500-message rollout. The pre-trained model and continued pre-training rise together, the selection null rises fastest, and post-training rises the slowest.}
    \caption{$\mathrm{KL}(\text{true}\,\Vert\,\text{gen})/H(\text{true})$ accumulated since each curve's first 25-message bin. Post-training accumulates $26\%$ less excess divergence than the anchor, the selection null $27\%$ more, continued pre-training the same. Shaded: 95\% day-clustered bootstrap bands.}
    \label{fig:compounding}
\end{figure}

We test the exposure-bias premise directly by measuring divergence as a function of continuation position. Over $4{,}096$ held-out context windows on the 28-day test panel, we compare the true continuation against a free-running rollout from the same context, one rollout per window, all rows generated in a single common-random-numbers program. At each token position we form the empirical marginal across windows over that position's valid sub-alphabet and compute $\mathrm{KL}(\text{true}\,\Vert\,\text{gen})$, normalized by the entropy of the true marginal to make positions of different difficulty comparable, and averaged over the sampled phases of the new-message block (two near-deterministic phases with vanishing entropy are excluded). Figure~\ref{fig:compounding} reports the accumulation view: each curve minus its own first 25-message bin, so every model starts at zero and its endpoint is the excess divergence accumulated over the rollout. Horizon-independent level offsets, including the estimator's finite-sample bias (a real-versus-real control of the same statistic is flat across the rollout), drop out of the comparison.

The pre-trained model's divergence grows $3.1\times$ from the first to the last 25-message bin, direct evidence that its errors compound over free-running generation. Post-training reduces exactly this accumulation: the ten-seed mean accumulates $26\%$ less excess divergence than the anchor over the 500-message horizon (paired day-bootstrap, replicate-wise differences on a draw matrix shared across rows; $95\%$ CI $[-33\%,-20\%]$). The controls locate the source of the reduction: the selection-luck null accumulates $27\%$ more than the anchor (CI $[+15\%,+37\%]$), and the continued-pre-training control accumulates at the anchor's rate ($+3.9\%$, CI $[-6\%,+14\%]$). Neither selection pressure nor additional likelihood training slows accumulation; the critic's sequence-level signal does.

\subsection{Market realism}

\subsubsection{Absolute value variations}
\label{sec:abs_variations}

\begin{figure}[h]
    \centering
    \includegraphics[width=\linewidth]{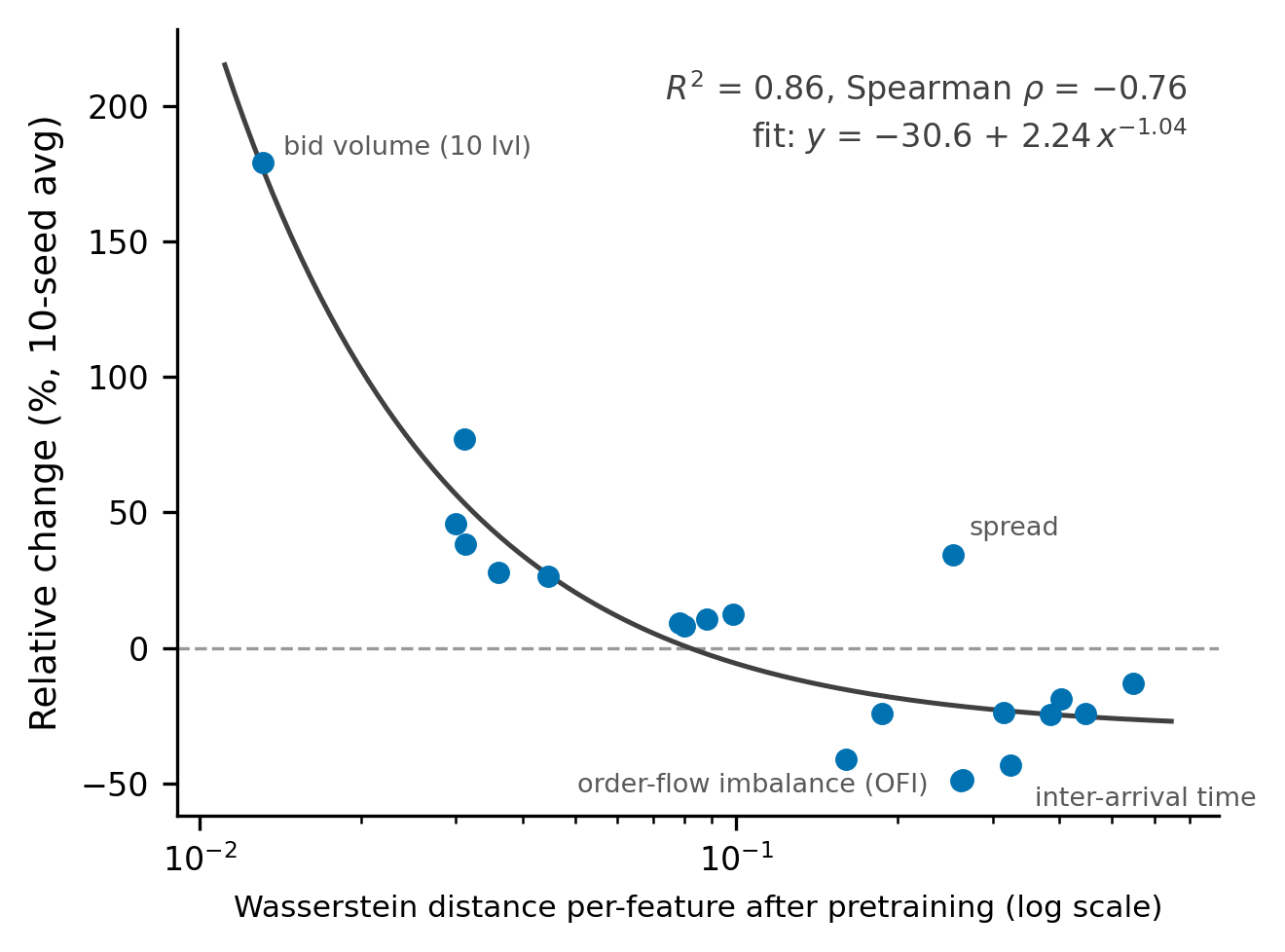}
    \Description{Scatter plot of relative error change after post-training against the pre-trained model's error for 21 features, with a decreasing power-law fit. Features with larger initial errors improve more; spread is marked as an outlier.}
    \caption{Per-feature error shift at test-time, averaged over the 10 post-training seeds. The saturating power-law fit ($R^2 = 0.86$, Spearman $\rho = -0.76$) shows the largest anchor errors improving most; features near the evaluation noise floor (left) swing widely in relative terms, and spread (annotated) is the main outlier.}
    \label{fig:abs_variations}
\end{figure}

Beyond the aggregate WS-21, per-feature response follows a regular pattern: the larger a feature's error in the base pre-trained model, the more post-training improves it. Fitting the 10-seed average relative change against the anchor's per-feature Wasserstein level with a saturating power law $y = y_0 + A\,x^{p}$ across the 21 features gives $R^2 = 0.86$ and Spearman $\rho = -0.76$ (Figure~\ref{fig:abs_variations}). The same compression holds across every critic variant of Section~\ref{sec:critic_ablations}, indicating it is a property of the adversarial signal itself rather than of any particular critic input.

The mechanism follows from the objective rather than from any feature. The critic scores whole rollouts and carries no per-feature term; its discriminative power concentrates wherever the anchor deviates most from the data, and that is where the gradient-free update is steered hardest. As those deviations shrink they stop separating real from generated and the pressure moves elsewhere---a saturating dynamic, which is why a saturating power law and not a linear trend describes the response. Because nothing in the objective identifies a feature, the pattern is a consequence of adversarial pressure seeking the cheapest separation, not feature-specific tuning.

Two departures are informative. Spread---despite the 9th-largest anchor error---worsens by $+34\%$. Unlike the compressed features it is not a marginal of the message stream but a joint property of the standing book, set by the interaction of placements and cancellations at the touch: the critic can improve each event type's marginal realism while the book-level equilibrium it induces drifts. It is also self-reinforcing under free running, since a slightly wide spread changes which orders are marketable, which changes the next message, which widens it further. Since the critic scores the message stream but never observes the standing book, this drift is only indirectly visible to the reward; the exchangeable reward module of Section~\ref{sec:future} offers a potential countermeasure, retargeting a follow-up post-training pass at book-level properties such as the spread itself. Conversely, features near the noise floor worsen only in relative terms: their anchor errors are so small that tiny absolute changes read as large percentages, and their absolute errors remain small after post-training.

\subsubsection{Beyond zero-sum shifts}

Recent work also characterizes the neighborhood of large pretrained models as a thicket of directions that improve individual metrics by chance~\cite{gan2026neural}, so checkpoint selection over a training run could in principle manufacture apparent gains from undirected drift alone. Here, it could be possible that the generator moves error between statistics---improving whatever the critic currently prices at the expense of what it ignores---without improving general performance. Our selection-null control bounds this channel by passing an equivalent number of random LoRA perturbations through the identical selection pipeline: it ends significantly above the anchor, so selection alone manufactures no gain here---likely because neural thickets are observed at far larger scales (beyond ${\sim}1.5$\,B parameters) than our $78.5$\,M-parameter generator.

\subsection{Downstream performance}
\label{sec:downstream}

Our results show that post-training improves market realism and slows error accumulation over free-running rollouts. We therefore ask if that gain also shows in downstream performance or if that improvement is compensated by worse predictive capabilities. We test this on mid-price direction: over $2{,}048$ held-out windows of the same 28-day panel, each model free-runs from a common context and we classify the sign of its mid-price change at horizons of $100$ to $1{,}000$ messages against the truth (three classes: up, flat, down). Deltas are paired against the anchor on identical windows, with a day-clustered bootstrap draw matrix shared across rows and horizons.

\begin{table}[h]
  \caption{Downstream mid-price direction at $2{,}048$ held-out windows (28-day panel). $\Delta$ is the mean change in three-class directional accuracy across $n{=}10$ seeds relative to the frozen pre-trained anchor, with paired day-clustered bootstrap $95\%$ CIs. Accuracy is preserved at every horizon: no interval separates from zero.}
  \label{tab:downstream}
  \begin{tabular}{@{}lr@{}}
    \toprule
    Horizon (messages) & $\Delta$ Directional accuracy [95\% CI] \\
    \midrule
    $100$     & $+0.011$\,\cis{-0.008}{+0.031} \\
    $250$     & $+0.012$\,\cis{-0.007}{+0.028} \\
    $500$     & $-0.004$\,\cis{-0.024}{+0.016} \\
    $1{,}000$ & $+0.009$\,\cis{-0.010}{+0.027} \\
    \bottomrule
  \end{tabular}
\end{table}

Directional accuracy is preserved at every horizon (Table~\ref{tab:downstream}). This is consistent with the two quantities living on different axes: realism concerns the distribution of order flow, while direction concerns the conditional mean of a near-martingale. Efficient prices leave almost no predictable component to sharpen (the anchor's own correlation with the true move decays to zero within ${\approx}400$ messages), so reshaping the flow distribution moves realism without disturbing direction.

Preservation is the informative outcome. No term in the critic or the ES reward targets directional prediction, so nothing protects this capability during the update: it is free to degrade, and does not. The compounding-error reduction is therefore not a zero-sum reallocation between capabilities. Rollouts stay closer to the data for longer (Section~\ref{sec:compounding}) while downstream predictive performance carries through intact, and that combination is what makes the refinement usable in practice.

\section{Conclusion}

We introduced an adversarial post-training framework for foundation models in high-frequency trading: a Wasserstein critic, trained online on rollouts executed through the true, non-differentiable matching engine, steers gradient-free updates, built from low-rank perturbations, of a pre-trained generator under a KL trust region. For under $50$\,M training tokens and $12$ GPU-hours per seed, three orders of magnitude below the pre-training token budget, post-training improves sealed held-out realism by $17.6\%$ (WS-21) and $17.0\%$ (Mean-$L_1$), every seed individually significant, and reduces the excess divergence accumulated over free-running 500-message rollouts by $26\%$. Selection-matched and compute-matched controls reproduce neither gain, isolating the critic's sequence-level signal as the active ingredient, and the improvement costs nothing downstream: directional accuracy is preserved at every horizon. Alongside the headline result, we characterized the reward-hacking window that bounds productive adversarial training, the selection protocol that disciplines it, and distilled ablations over critic inputs, optimizers, and population geometry into reusable design guidelines.

\subsection{Future work}
\label{sec:future}
\label{sec:generalization}

Because nothing in the pipeline is specific to aggregate realism---the critic is an exchangeable reward module---our framework transfers to any property computable from free rollouts, inside or outside the high-frequency literature. Post-training thus forms its own stage between pre-training and fine-tuning: pre-train once for global market structure, post-train to sharpen a targeted property, and fine-tune the result into a specialized system such as an agentic trading framework or an information-decoding pipeline. The most consequential target is market-impact fidelity: empirical impact obeys tight regularities such as the square-root law~\cite{almgren2005direct,bouchaud2009digest}, and a simulator that reproduces the counterfactual price response to injected orders is the missing prerequisite for training and evaluating trading agents in realistic environments~\cite{byrd2020abides,frey2023jaxlob}. Another idea could be to adapt our method towards different objectives: a GAN discriminator could, for example, post-train a variant of our model in order to perform market making, learning to score policies based on expected profits. 

Two further directions concern the pipeline itself. First, all rollouts here span $500$ messages; whether longer training horizons slow error accumulation further remains open. Second, the trained critic is a realism score in its own right: used in isolation, it could rank generators, flag anomalous order flow, or act as a frozen reward, enabling a controlled study of reward hacking apart from the co-adaptation of Section~\ref{sec:hacking}.

\begin{acks}
    This work was supported by the Engineering and Physical Sciences Research Council [grant number UKRI5149] and other \href{https://www.bold-lab.ai/sponsors}{sponsors of BOLD}. The authors acknowledge the use of resources provided by the Isambard-AI National AI Research Resource (AIRR). Isambard-AI is operated by the University of Bristol and is funded by the UK Government’s Department for Business, Innovation, Science and Trade (DBIST) via UK Research and Innovation; and the Science and Technology Facilities Council [ST/AIRR/I-A-I/1023]~\cite{mcintoshsmith2024isambardaileadershipclasssupercomputer}. 
    
    AB gratefully acknowledges the support of the Anne Dias Griffin and
    Kenneth C. Griffin Harvard University Fellowship. VM gratefully acknowledges funding from the Department of Computer Science at the University of Oxford. AC acknowledges funding from a UKRI AI World Leading Researcher Fellowship (grant EP/W002949/1). AC and JF acknowledge funding from a JPMC Faculty Research Award.
\end{acks}

\balance 

\bibliographystyle{ACM-Reference-Format}
\bibliography{references}

\end{document}